# Overview of the HASOC track at FIRE 2020: Hate Speech and Offensive Content Identification in Indo-European Languages

Thomas Mandl[a,g], Sandip Modha[b], Gautam Kishore Shahi[c], Amit Kumar Jaiswal[d,h], Durgesh Nandini[e], Daksh Patel[f], Prasenjit Majumder[g] and Johannes Schäfer[a]

[a]*University of Hildesheim, Germany*
[b]*LDRP-ITR, Gandhinagar, India*
[c]*University of Duisburg-Essen, Germany*
[d]*University of Bedfordshire, United Kingdom*
[h]*University of Leeds, United Kingdom*
[e]*University of Bamberg, Germany*
[f]*Dalhousie University, Halifax, Canada*
[g]*DA-IICT, Gandhinagar, India*

**Abstract**
With the growth of social media, the spread of hate speech is also increasing rapidly. Social media are widely used in many countries. Also Hate Speech is spreading in these countries. This brings a need for multilingual Hate Speech detection algorithms. Much research in this area is dedicated to English at the moment. The HASOC track intends to provide a platform to develop and optimize Hate Speech detection algorithms for Hindi, German and English. The dataset is collected from a Twitter archive and pre-classified by a machine learning system. HASOC has two sub-task for all three languages: task A is a binary classification problem (Hate and Not Offensive) while task B is a fine-grained classification problem for three classes (HATE) Hate speech, OFFENSIVE and PROFANITY. Overall, 252 runs were submitted by 40 teams. The performance of the best classification algorithms for task A are F1 measures of 0.51, 0.53 and 0.52 for English, Hindi, and German, respectively. For task B, the best classification algorithms achieved F1 measures of 0.26, 0.33 and 0.29 for English, Hindi, and German, respectively. This article presents the tasks and the data development as well as the results. The best performing algorithms were mainly variants of the transformer architecture BERT. However, also other systems were applied with good success.

# 1. Introduction: Hate Speech and Its Identification

The large quantity of posts on social media has led to a growth in problematic content. Such content is often considered harmful for a rationale and constructive debate [1]. Many countries have defined more and more detailed rules for dealing with offensive posts [2, 3]. Companies and platforms are also concerned about problematic content. Problematic content and in particular hate speech has been a growing research area. Linguists have analysed and described various forms of hate speech [4]. Political scientists and legal experts search for ways to regulate platforms and to handle problematic content without oppressing free speech [5].

The identification of hate speech within large collections of posts has led to much research in information science and computer science. Much research is carried out within big internet platforms. It is important to provide open resources to keep the society informed about the current performance of technology and the challenges of hate speech identification.

Algorithms are continuously improved and diverse collections for a variety of related tasks and for several languages are being generated and analysed. Collections were built recently for many languages [6], e.g. for Greek, [7], Portuguese [8], Danish [9], Mexican Spanish [10] and Turkish [11]. The availability of several benchmarks allows better analysis of their differences and their reliability [12].

HASOC contributes to this research, in this second edition, a different approach for creating the data was applied. The two main tasks and the languages were kept identical for better comparison with the results from HASOC 2019 [13].

# 2. HASOC Task Description

In HASOC 2020, two tasks in the research area of Hate Speech detection are proposed. These two sub-tasks are offered for all three languages: Hindi, German and English. We chose in particular Hindi and German as languages fewer less resources. HASOC also provides a testbed for English to see how the algorithms perform in comparison to a language with many resources. Below is a description of each task.

## 2.1. Sub-task A: Identifying Hate-Offensive and Non Hate-Offensive content (Binary)

This task focuses on Hate speech and Offensive language identification offered for English, German, and Hindi. Sub-task A is coarse-grained binary classification in which participating system are required to classify tweets into two classes, namely: Hate and Offensive (HOF) and Non- Hate and offensive (NOT).

- **NOT - Non Hate-Offensive**: This post does not contain any Hate speech, profane, offensive content.

- **HOF - Hate and Offensive**: This post contains Hate, offensive, and profane content.

| | | |
|---|---|---|
| RT @rjcmxrell: im not fine, i need you | NOT | NONE |
| You be playin= I'm tryna fuck | HOF | PRFN |
| RT @femmevillain: jon snow a punk ass bitch catelyn was right to bully him | HOF | OFFN |
| Buhari His Not Our President, I'm Ready To Go To Prison – Ayo Adebanjo Dares FG https://t.co/XXR6VRRI5b | NOT | NONE |
| This shit sad af "but I don't have a daddy" ... you niggas gotta do better by these kids they didn't ask to be here. | HOF | HATE |
| RT @GuitarMoog: As for bullshit being put about by people who do know better, neither of the two biggest groups in the EP are going to get… | HOF | PRFN |

**Table 1**
Examples for each class from the data set

## 2.2. Sub-task B: Identifying Hate, Profane and Offensive posts (fine-grained)

This sub-task is a fine-grained classification also offered for English, German, and Hindi. Hate-speech and offensive posts from the sub-task A are further classified into three categories:

- **HATE - Hate speech**: Posts under this class contain Hate speech content. Ascribing negative attributes or deficiencies to groups of individuals because they are members of a group (e.g. all poor people are stupid) would belong to this class. In more detail, this class combines any hateful comment toward groups because of race, political opinion, sexual orientation, gender, social status, health condition or similar.

- **OFFN - Offensive**: Posts under this class contain offensive content. In particular, this refers to degrading, dehumanizing or insulting an individual. Threatening with violent acts also belongs to this class.

- **PRFN - Profane**: These posts contain profane words. Unacceptable language in the absence of hate and offensive content. This typically concerns the usage of obscenity, swearwords and cursing.

Some examples for posts from all classes from the final set are shown in Table 1.

## 3. Dataset Description

Most hate speech datasets, including HASOC 2019 [13], are sampled by crawling social media platforms or addressing their API using keywords considered relevant for hate speech or scrapping hashtags. As a variant, these methods are used to find user of social media who frequently posts hate speech message and collect all message from their timeline [14]. All of these methods are based on hand crafted lists of hate speech related terms. This may introduce a bias because this process might limit the collection to topics and word which are remembered. An empirical analysis [15] has pointed out that these practices may lead to bias. It concluded that datasets that are created using focused sampling exhibit more bias than those crated by random sampling. Furthermore, deep learning or machine learning models that reported the

| Class | English | German | Hindi |
|---|---|---|---|
| NOT | 1,852 | 1,700 | 2,116 |
| HOF | 1,856 | 673 | 847 |
| *PRFN* | *1,377* | *387* | *148* |
| *HATE* | *158* | *146* | *234* |
| *OFFN* | *321* | *140* | *465* |
| Sum | 3,708 | 2,373 | 2,963 |

**Table 2**
Statistical overview of the Training Data

| Class | English | German | Hindi |
|---|---|---|---|
| NOT | 391 | 392 | 466 |
| HOF | 423 | 134 | 197 |
| *PRFN* | *293* | *88* | *27* |
| *HATE* | *25* | *24* | *56* |
| *OFFN* | *82* | *36* | *87* |
| Sum | 814 | 526 | 663 |

**Table 3**
Statistical overview of the Development Data for providing performance measures during submission

| Class | English | German | Hindi |
|---|---|---|---|
| NOT | 395 | 393 | 436 |
| HOF | 418 | 133 | 226 |
| *PRFN* | *287* | *89* | *36* |
| *HATE* | *33* | *33* | *57* |
| *OFFN* | *87* | *30* | *104* |
| Sum | 813 | 526 | 662 |

**Table 4**
Statistical overview of the Test Data for determining the final results best results on these biased datasets substantially underperform on benchmark datasets sampled using different keywords. Similar observations on bias induced from training data were also reported [16]. However, fully random sampling leads to dataset with very few hate speech samples which requires much more manual annotation.

One of the HASOC 2020 dataset's main objectives is to minimize the impact of bias in the training data. The observations made during previous research Wiegand et al. [15], Davidson et al. [16] inspired us to develop a hate speech dataset based on a sampling process which relies on less input. The final size of the training, development and testing sets are shown in Tables 2, 3 and 4. The development set was used to calculate the metrics for participants during the campaign.

During planning for HASOC 2020, the organizers searched for tweets collections and iden-tified archive.org[1]. We have downloaded the entire archive for month May 2019. The archive contains tweets on an hourly basis. The volume of the tweets for a full day is around 2.25 GB in compressed format and 22.9 GB in uncompressed format. To obtain a set of hateful tweet, we developed a sampling method that will be presented in the next paragraph.

After downloading the archive of tweets, we extracted English, German, and Hindi tweets using the language attribute provided by the Twitter metadata. We trained a SVM machine learning model with word features weighted by TF-IDF without considering any further features.

For the dataset development, the entire May 2019 archive was crawled for German. For English and Hindi the archives of 1st, 10th and 19th May 2019 were crawled. We used Python scripts for filtering.

We found an average of some 1301000 tweets for English from the archive for each day. We sampled only 35000 tweets as potential hate speech candidates. Similarly, the average volume of Hindi tweets is around 24000 tweets and the amount of German tweets is around 12500.

To obtain potentially hateful tweets in English, we have trained the SVM model on the OLID [17] and HASOC 2019 dataset. The purpose was to create a weak binary classifier that gives an F1-score around 0.5. We have tested these models on the English tweets that were extracted from the archive. We considered all the tweets that are classified as hateful by the week classifier. We added 5 percent of the tweets which were not classified as hateful randomly. The main idea behind this merge is to ensure that the final dataset contains an equal distribution of hateful and non-hateful tweets. Then this set of English tweets was distributed to the annotators using heuristics within the platform. Out of 35000 tweets, some 2600 tweets were labeled as potentially hateful tweets by the classifier.

The Hindi dataset was prepared using the same method, but the SVM model was trained with the TRAC corpus [18] and the HASOC 2019 corpus. The average total no of the potential hate speech around 5700 out of 24000(around 24 percent). The German dataset was extracted from the archive using the SVM model that was trained with the dataset from GERMeval 2018 [19] [2] and the dataset from HASOC 2019. It is worth to note that the number of the potentially hateful tweets found for in English and Hindi is substantially higher than for German (more than ten times). Therefore, we had to crawl the entire month May 2019 to obtain a dataset of reasonable size for German. The weak classifier based on a SVM labeled only 150 (around 1.25 percent) tweets as a potentially hateful tweets.

### 3.1. Data Annotation

All tweets in these sets were annotated manually by people who use social media in the respective language. The annotators were students who received a small amount of money for their work. They were neither aware how the tweets were collected nor did they know about the classification result of a tweet.

**Tweet allocation for annotations**  The tweet allocation was performed in such a way that each tweet was annotated twice. In case when there was a conflict in the annotation between the first two annotators, the tweet is automatically scheduled to be assigned to a third annotator who had not yet seen the respective tweet. This way we ensured the integrity of the annotation, and try to avoid human bias. Annotators could also report tweets for a variety of reasons. In cases a particular tweet was reported by both the initial annotators, then it is considered as an outlier and not used further while generating the dataset. However, the resources for labelling were limited, so not all conflict cases could be resolved by a third annotation.

---

[1]https://archive.org/details/archiveteam-twitter-stream-2019-05
[2]https://projects.cai. fbi.h-da.de/iggsa/

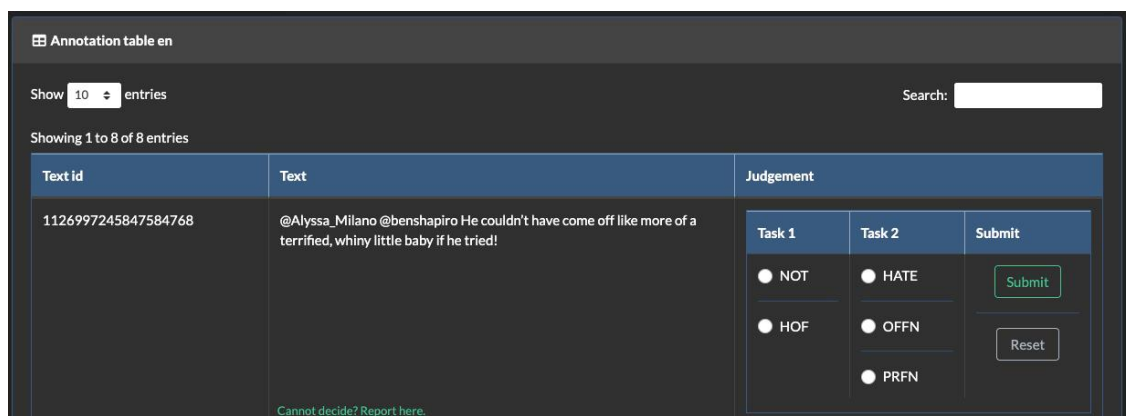

**Figure 1:** Screenshot of annotation interface: Tweet View

**Annotation Platform and Process** During the labelling process, the annotators for each language engaged with an online system to judge the tweets. The online system presented the text of the tweet only and users had to make the decision. The annotation system allows the oversight of the process so that progress can be monitored.

The interface of the system can be seen in Figure 1 and Figure 2. For Hindi and German, native speakers were contracted as annotators. For English, students from India (Gujarat) were contracted who are educated in English and who use social media regularly in English. The annotators were given short guidelines that contained the information as mentioned in section 2.1 and 2.2. Apart from the definitions listed above, the guidelines listed the following rules.

- Dubious cases which are difficult to decide even for humans, should be left out.
- Content behind links is not considered
- Hashtags are considered
- Content in other languages is not labelled, but notified

The annotators met online in brief meetings during the process at least twice for each language. They discussed the guidelines and borderline cases in order to find a common practice and interpretation of hate speech.

Nevertheless, the process remains highly subjective, and even after discussions of questionable often no agreement could be reached. This lies in the nature of Hate Speech and its perception by humans. Overall, the new sampling method led to a large portion of profane content.

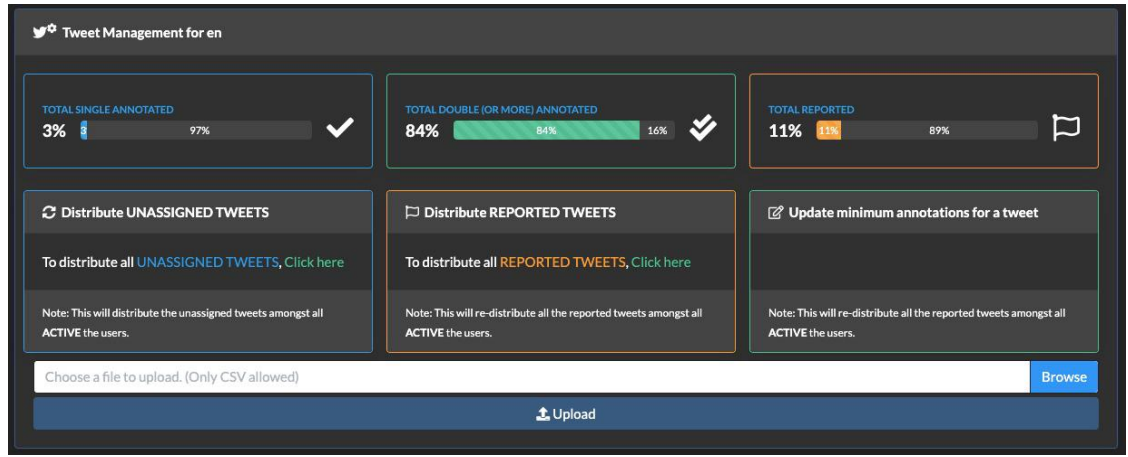

**Figure 2:** Screenshot of Annotation Interface: Management View

| Language | Task | Agreement(%) |
|---|---|---|
| English | Task A | 82.7 |
| | Task B | 69.2 |
| Hindi | Task A | 65.4 |
| | Task B | 56.4 |
| German | Task A | 83.3 |
| | Task B | 78.7 |

**Table 5**
Inter-coder reliability of different task across different language

### 3.2. Inter-Coder Reliability

We randomly assigned the data to the annotators, and two or three annotators annotated each tweet at a time. For tweets with three annotators the majority vote was considered. For the tweets annotated by two annotators, we used these two approaches, case I: When both annotators voted for the same judgment, this majority decision is accepted. Case II: Two annotators contradict each other, and we considered the rating of the more reliable annotator. We measured the reliability of the annotators with heuristics based on the overlap of their previous annotations with others. The data were annotated by 11, 11 and 8 different annotators for Hindi, English, and German.

The first round data was annotated by different annotators, and the majority vote was considered, the annotation agreement is shown in table 5. For the disagreed data, we considered the second round of annotation. For the English and German language, we considered the voting of the most reliable voters. The algorithm used for determining the voting reliability of each annotator is shown in figure 3. For Hindi, we re-annotated the conflicted data with different annotators.

```
Algorithm 1 To assign the best voting
 1: procedure ANNOTATORS(labelleddata)▷ ranking the annotator's reliability
 2:     annotators ← [List of annotators]
 3:     totalvoting ← [Number voting done by each annotator]
 4:     majorityvote ← [Count the majority vote for each records])
 5:     assign_score ← [Assign 1 if the the majority vote is same of final vote else 0]
 6:     averagescore ← assign_score      ▷ take the average of all voting done by each annotator
 7:     ranking ← averagescore                            ▷ Ranking of all annotators
 8:     if conflict in annotation then
 9:         annotator ← best_annotator                      ▷ Pick the best annotator
10:         label ← best_voting                            ▷ Voting of best_annotator
11:     return label
```

**Figure 3:** Data annotation based on reliability of annotators

## 4. Participation and Evaluation

This section details the participants submission and the evaluation method in each of the three language sub-tasks i.e., English, German and Hindi. Each language tasks consist of 2 sub-tasks and registered teams were able to take part in any of the sub-tasks respectively. There were 116 registered participants and 40 teams finally submitted results.

### 4.1. Submission Format

The submission and evaluation of experiments was handled on Codalab [3]. The system can be seen in figure 4.

### 4.2. Performance Measure

We posed two sub-tasks for each of the languages - English, German and Hindi. As each of the language sub-tasks contains multiple classes with non-uniform numbers of samples, we decided to use an item weighted measure to rank the submission of the teams, in our case the *macro F1* measure.

### 4.3. Evaluation Timeline

The participants could receive online access to the training data and worked on the tasks. The data of HASOC 2019 was also available for participants, however, not many reported using it. During that phase, they could observe their performance based on the development set. In particular, the position relative to other teams is interesting.

- Release of Training data: September 15, 2020

---
[3]https://competitions.codalab.org/competitions/26027

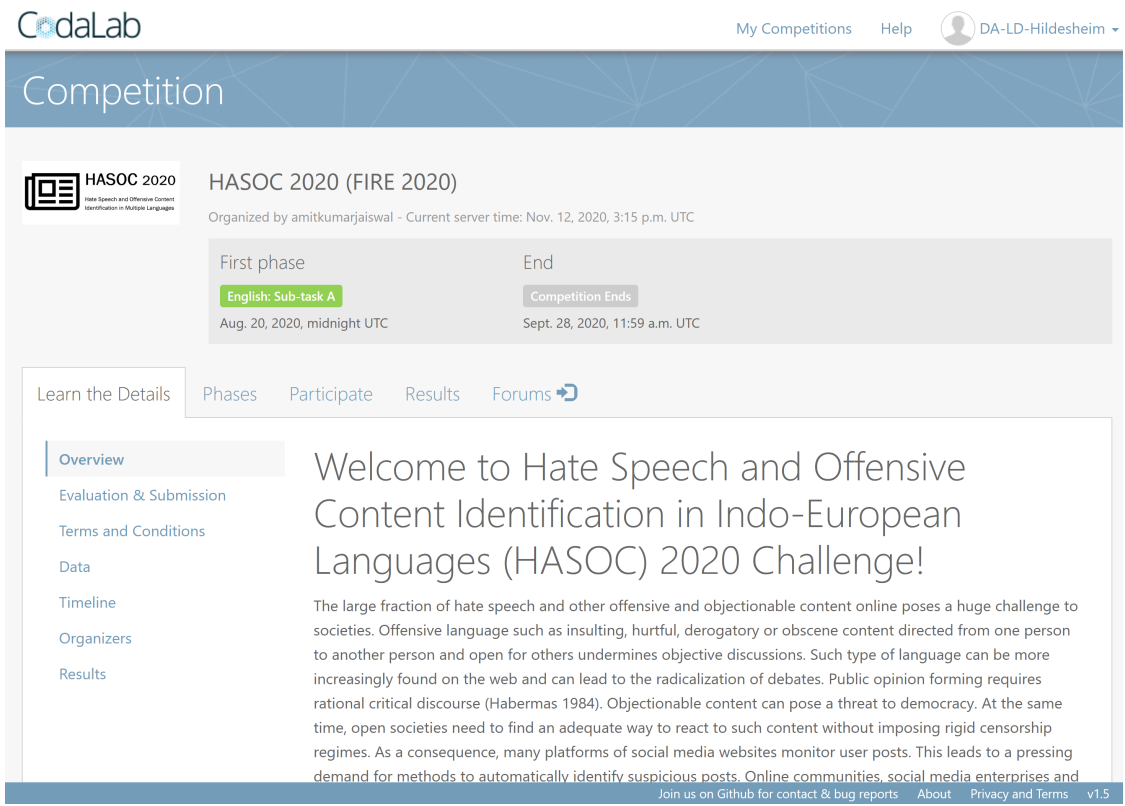

**Figure 4:** Screenshot of HASOC Codablab Website

- Result submission on Codalab: September 27, 2020

Overall, more than 252 experiments were submitted.

## 5. Results for Tasks

This section gives the results of the participating systems. They are ordered by language and each subsection reports of both tasks. Unfortunately, a description was not submitted for all systems. All metrics in the following tables are reported for the test set 4. The results overall prove that the task of identifying and further classifying hate speech of offensive language is still challenging. No F1 score above 0.55 could be achieved. These scores are lower than those achieved for the HASOC 2019 dataset [13].

### 5.1. Hindi

The results for Hindi are shown in Tables 6 and 7. The submission data shows the performance and the date of submission in Figures 5 and 6. It suggests that the leaderboard on the website was helpful for some teams to improve their score.

| Rank | Team Name | Entries | F1 Macro average |
|------|-----------|---------|------------------|
| 1 | NSIT_ML_Geeks | 1 | 0.5337 |
| 2 | Siva | 1 | 0.5335 |
| 3 | DLRG | 2 | 0.5325 |
| 4 | NITP-AI-NLP | 1 | 0.5300 |
| 5 | YUN111 | 1 | 0.5216 |
| 6 | YNU_OXZ | 2 | 0.5200 |
| 7 | ComMA | 4 | 0.5197 |
| 8 | Fazlourrahman Balouchzahi | 3 | 0.5182 |
| 9 | HASOCOne | 1 | 0.5150 |
| 10 | HateDetectors | 2 | 0.5129 |
| 11 | IIIT_DWD | 1 | 0.5121 |
| 12 | LoneWolf | 2 | 0.5095 |
| 13 | MUM | 2 | 0.5033 |
| 14 | IRLab@IITVaranasi | 2 | 0.5028 |
| 15 | CONCORDIA_CIT_TEAM | 1 | 0.5027 |
| 16 | QutBird | 2 | 0.4992 |
| 17 | Oreo | 2 | 0.4943 |
| 18 | CFILT IIT Bombay | 1 | 0.4834 |
| 19 | TU Berlin | 1 | 0.4678 |
| 20 | JU | 1 | 0.4599 |

Table 6
Results Task A Hindi: Top 20 Submission

The best result for task A received an F1 score of slightly above 0.53. This can be considered a low score and the tasks proved to be more challenging than the HASOC 2019 experiments. The 10 best submissions received very similar scores.

The best system applied a BiLSTM with 1 layer and fastText word embeddings as basic representation for the input [20]. The submission at position 3 has not used any deep learning but a lexical approach with TF-IDF weighting and a SVM for classification. This system also translated the tweets automatically to augment the set of available training samples[21]. The fourth system has applied the BERT model distilBERT-base for different languages [22].

For task B, the performance is even lower and the best experiment reaches only a score above 0.33. However, the first system has a much better performance than the following submissions. Rank 2 to 10 score again very similar.

For the best ranked system, uses fine-tuned BERT model for the classification [23]. The second-ranked system was already successful for task A. It applied a BiLSTM and fastText as basic representation[20]. The third ranked team from LDRP-ITR experimented with BERT and GPT-2. For this run, they used a CNN which conducted a bigram and a trigram analysis in parallel and fused the results [24].

| Rank | Team Name | Entries | F1 Macro average |
|------|-----------|---------|------------------|
| 1 | Sushma Kumari | 1 | 0.3345 |
| 2 | NSIT_ML_Geeks | 1 | 0.2667 |
| 3 | Astralis | 1 | 0.2644 |
| 4 | Oreo | 1 | 0.2612 |
| 5 | Siva | 1 | 0.2602 |
| 6 | HASOCOne | 2 | 0.2574 |
| 7 | MUM | 3 | 0.2488 |
| 8 | ComMA | 5 | 0.2464 |
| 9 | AI_ML_NIT_Patna | 1 | 0.2399 |
| 10 | IIIT_DWD | 1 | 0.2374 |
| 11 | CFILT IIT Bombay | 1 | 0.2355 |
| 12 | CONCORDIA_CIT_TEAM | 1 | 0.2323 |
| 13 | HateDetectors | 1 | 0.2272 |
| 14 | YUN111 | 1 | 0.2100 |
| 15 | SSN_NLP_MLRG | 2 | 0.2063 |

**Table 7**
Results Task B Hindi: Top 15 Submission

### 5.2. German

The results for German are shown in Tables 8 and 9. The submission data is given in Figures 7 and 8. The situation for German is similar to the results of Hindi. The best F1 score is not very high and the best submissions are close to each other.

The best performance for German was achieved using fine-tuned versions of BERT, DistilBERT and RoBERTa [25]. Also the second best system used BERT. The group adapted the upper layer structure of BERT-Ger [26]. Also other systems have applied BERT and variants, e.g. position 4 [27], position 8 [22] and position 14 [28].

For task B, the results are very close together. The best model was submitted by team Siva [29]. The second best submission used ALBERT [28]. For the third rank, experiments with versions of BERT, DistilBERT and RoBERTa were submitted [25]. Huiping Shi used a self-attention model [30].

### 5.3. English

English attracted most experiments for both tasks. The results for English are shown in tables 10 and 11. Again, submission data is summarized in Figures 9 and 10. The performance differences between the top 30 teams are extremely small. The relative improvement is about 5%. Like for Hindi and German, the F1-measure shows rather low values. The best system achieved a performance of of 0.52.

The best result for English is based on a LSTM which used GloVe embeddings as input [31]. The TU Berlin team used a character based LSTM which performed better than their experiments with BERT [32]. Many other submissions used BERT. The team in position 6 has used a self-developed transformer architecture [30]. The team from IIT Patna used a standard BERT model and reached the third position [33]. The team from Jadavpur University (JU) [34] and one

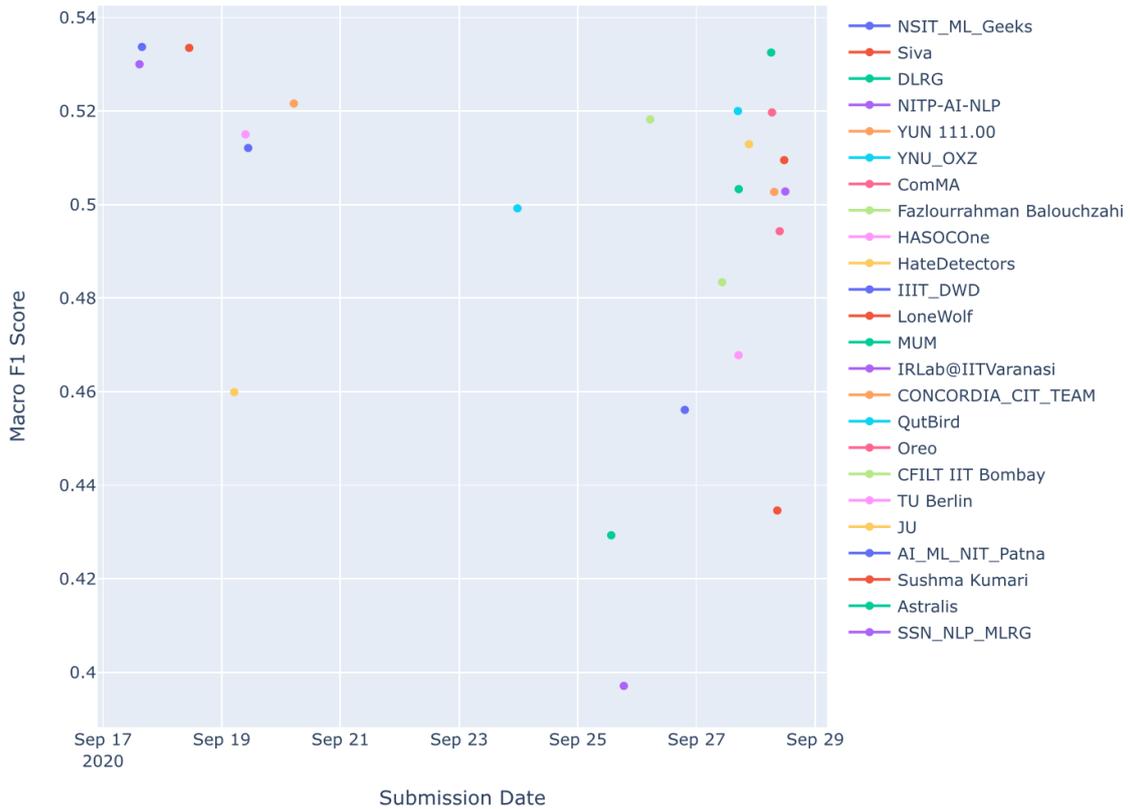

**Figure 5:** Hindi Task A

team from Yunnan University [27] used RoBERTa. The second team from Yunnan University applied a ensemble of three classifiers including BERT, LSTM and CNN [35].

For task B, the top three systems used BERT and variants. The best result was achieved by team Chrestotes with a F1 value of 0.26 for English [36]. They used a fine-tuned version of BERT. The team HUB from Yunnan University applied ALBERT and BERT. The team ZEUS from Yunnan University applied ALBERT and DPCNN [37].

### 5.4. Leaderboard

We report the participants statistics at team-level across all of the three languages for the corresponding sub-tasks in Table 12.

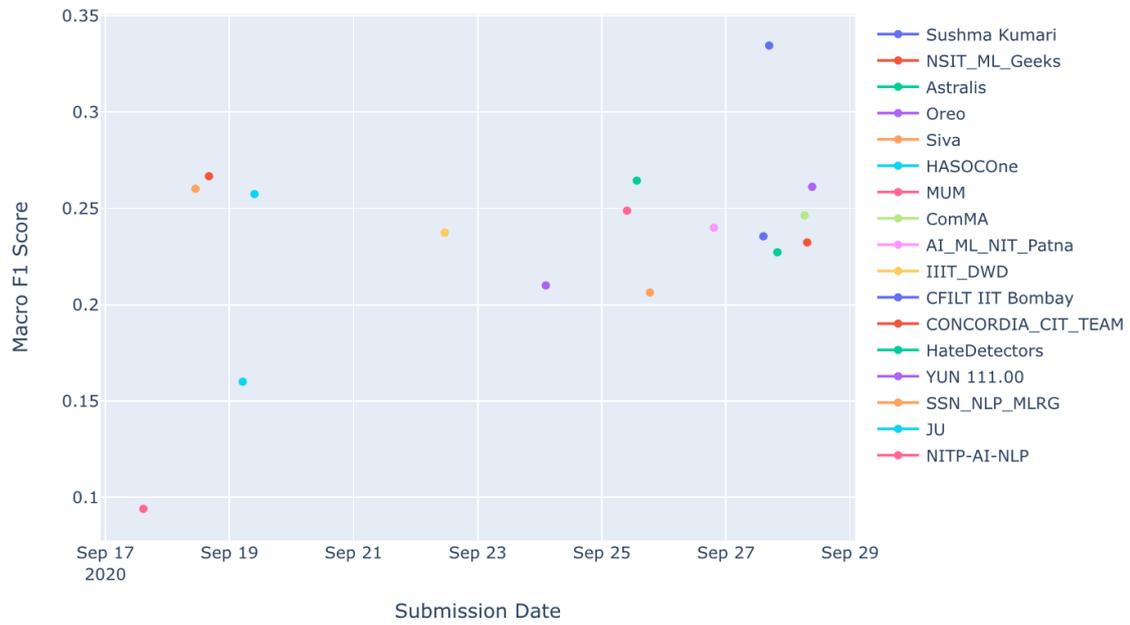

**Figure 6:** Hindi Task B

| Rank | Team Name | Entries | F1 Macro average |
|---|---|---|---|
| 1 | ComMA | 4 | 0.5235 |
| 2 | simon | 1 | 0.5225 |
| 3 | CONCORDIA_CIT_TEAM | 1 | 0.5200 |
| 4 | YNU_OXZ | 3 | 0.5177 |
| 5 | Siva | 1 | 0.5158 |
| 6 | Buddi_avengers | 2 | 0.5121 |
| 7 | Huiping Shi | 2 | 0.5121 |
| 8 | NITP-AI-NLP | 1 | 0.5109 |
| 9 | MUM | 1 | 0.5106 |
| 10 | HASOCOne | 4 | 0.5054 |
| 11 | Fazlourrahman Balouchzahi | 2 | 0.5044 |
| 12 | Oreo | 1 | 0.5036 |
| 13 | CFILT IIT Bombay | 1 | 0.5028 |
| 14 | SSN_NLP_MLRG | 2 | 0.5025 |
| 15 | IIIT_DWD | 1 | 0.5019 |
| 16 | yasuo | 1 | 0.4968 |
| 17 | hub | 2 | 0.4953 |
| 18 | NSIT_ML_Geeks | 2 | 0.4919 |
| 19 | DLRG | 2 | 0.4843 |
| 20 | Astralis | 1 | 0.4789 |

**Table 8**
Results Task A German: Top 20 Submissions

| #  | Team Name          | Entries | F1 Macro average |
|----|--------------------|---------|------------------|
| 1  | Siva               | 1       | 0.2943           |
| 2  | SSN_NLP_MLRG       | 1       | 0.2920           |
| 3  | ComMA              | 4       | 0.2831           |
| 4  | Huiping Shi        | 1       | 0.2736           |
| 5  | CONCORDIA_CIT_TEAM | 1       | 0.2727           |
| 6  | Astralis           | 1       | 0.2627           |
| 7  | Buddi_avengers     | 2       | 0.2609           |
| 8  | MUM                | 2       | 0.2595           |
| 9  | CFILT IIT Bombay   | 1       | 0.2594           |
| 10 | simon              | 1       | 0.2579           |
| 11 | hub                | 1       | 0.2567           |
| 12 | Oreo               | 1       | 0.2542           |
| 13 | IIIT_DWD           | 1       | 0.2513           |
| 14 | NSIT_ML_Geeks      | 2       | 0.2468           |
| 15 | HASOCOne           | 4       | 0.2397           |

**Table 9**
Results Task B German: Top 15 Submissions

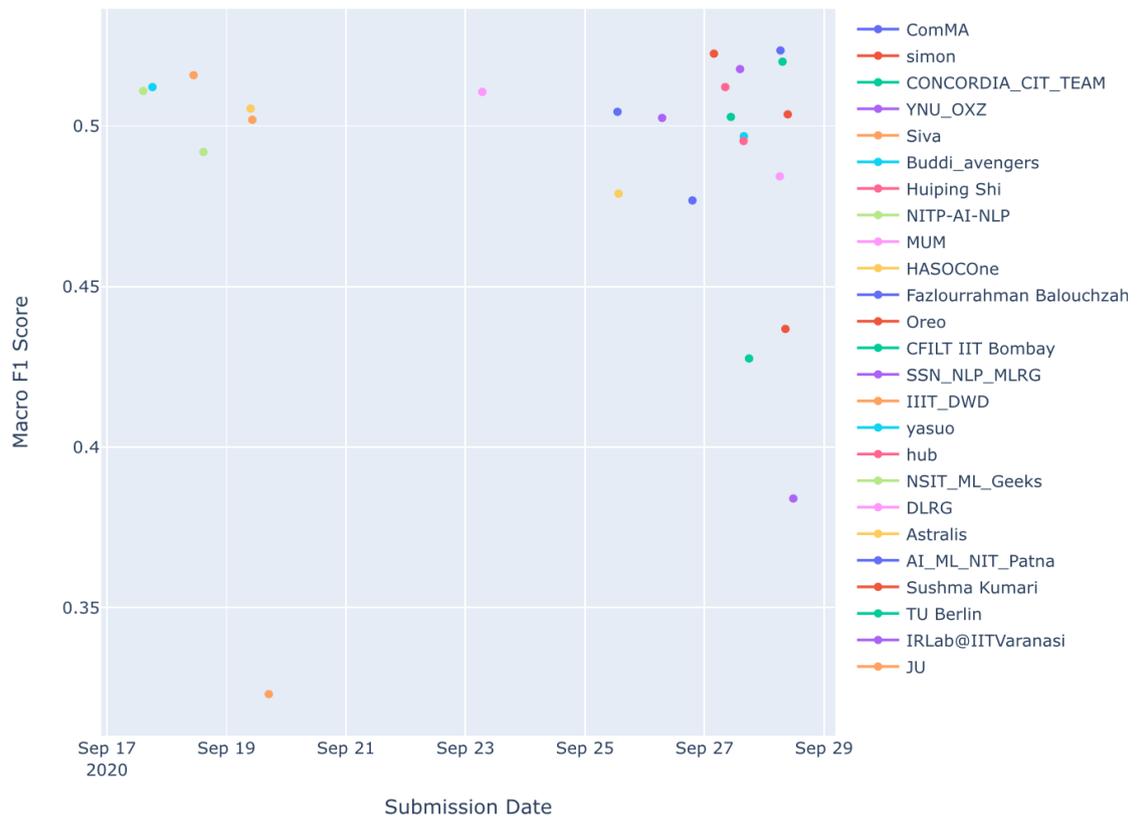

**Figure 7:** German Task A

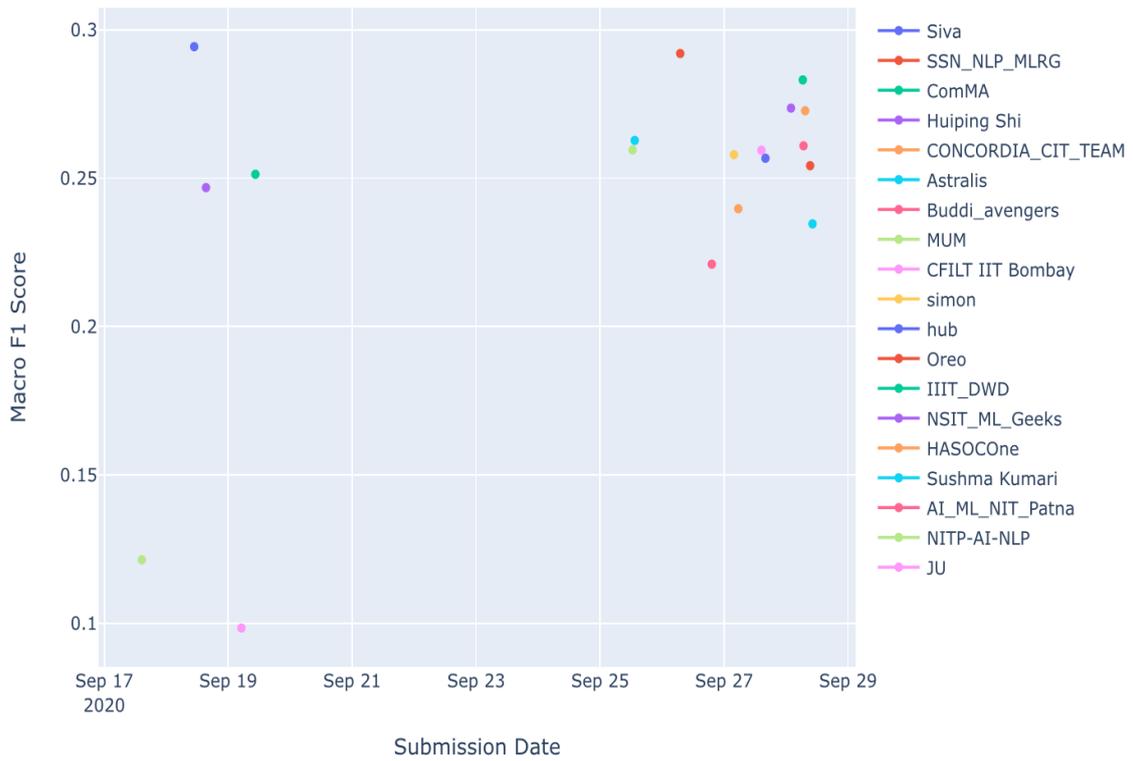

**Figure 8:** German Task B

| Rank | Team Name | Entries | F1 Macro average |
|---|---|---|---|
| 1 | IIIT_DWD | 1 | 0.5152 |
| 2 | CONCORDIA_CIT_TEAM | 1 | 0.5078 |
| 3 | AI_ML_NIT_Patna | 1 | 0.5078 |
| 4 | Oreo | 6 | 0.5067 |
| 5 | MUM | 3 | 0.5046 |
| 6 | Huiping Shi | 6 | 0.5042 |
| 7 | TU Berlin | 1 | 0.5041 |
| 8 | NITP-AI-NLP | 1 | 0.5031 |
| 9 | JU | 2 | 0.5028 |
| 10 | HASOCOne | 6 | 0.5018 |
| 11 | Astralis | 2 | 0.5017 |
| 12 | YNU_WU | 3 | 0.5017 |
| 13 | YNU_OXZ | 2 | 0.5006 |
| 14 | HRS-TECHIE | 6 | 0.5002 |
| 15 | ZYJ | 2 | 0.4994 |
| 16 | Buddi_SAP | 2 | 0.4991 |
| 17 | HateDetectors | 2 | 0.4981 |
| 18 | QutBird | 8 | 0.4981 |
| 19 | NLP-CIC | 2 | 0.4980 |
| 20 | SSN_NLP_MLRG | 1 | 0.4979 |
| 21 | Fazlourrahman Balouchzahi | 4 | 0.4979 |
| 22 | Lee | 1 | 0.4976 |
| 23 | IRIT-PREVISION | 2 | 0.4969 |
| 24 | chrestotes | 1 | 0.4969 |
| 25 | zeus | 1 | 0.4954 |
| 26 | DLRG | 4 | 0.4951 |
| 27 | ComMA | 4 | 0.4945 |
| 28 | Siva | 1 | 0.4927 |
| 29 | hub | 2 | 0.4917 |
| 30 | CFILT IIT Bombay | 2 | 0.4889 |

**Table 10**
Results Task A English: Top 30 Submissions

| Rank | Team Name | Entries | F1 Macro average |
|---|---|---|---|
| 1 | chrestotes | 2 | 0.2652 |
| 2 | hub | 1 | 0.2649 |
| 3 | zeus | 1 | 0.2619 |
| 4 | Oreo | 2 | 0.2529 |
| 5 | Fazlourrahman Balouchzahi | 4 | 0.2517 |
| 6 | Astralis | 1 | 0.2484 |
| 7 | QutBird | 1 | 0.2450 |
| 8 | Siva | 1 | 0.2432 |
| 9 | Buddi_SAP | 2 | 0.2427 |
| 10 | HRS-TECHIE | 4 | 0.2426 |
| 11 | ZYJ | 1 | 0.2412 |
| 12 | ComMA | 4 | 0.2398 |
| 13 | Huiping Shi | 5 | 0.2396 |
| 14 | Buddi_avengers | 1 | 0.2391 |
| 15 | MUM | 2 | 0.2388 |
| 16 | NSIT_ML_Geeks | 1 | 0.2361 |
| 17 | HASOCOne | 7 | 0.2357 |
| 18 | IIIT_DWD | 1 | 0.2341 |
| 19 | SSN_NLP_MLRG | 1 | 0.2305 |
| 20 | HateDetectors | 2 | 0.2299 |

**Table 11**
Results Task B English: Top 20 Submissions

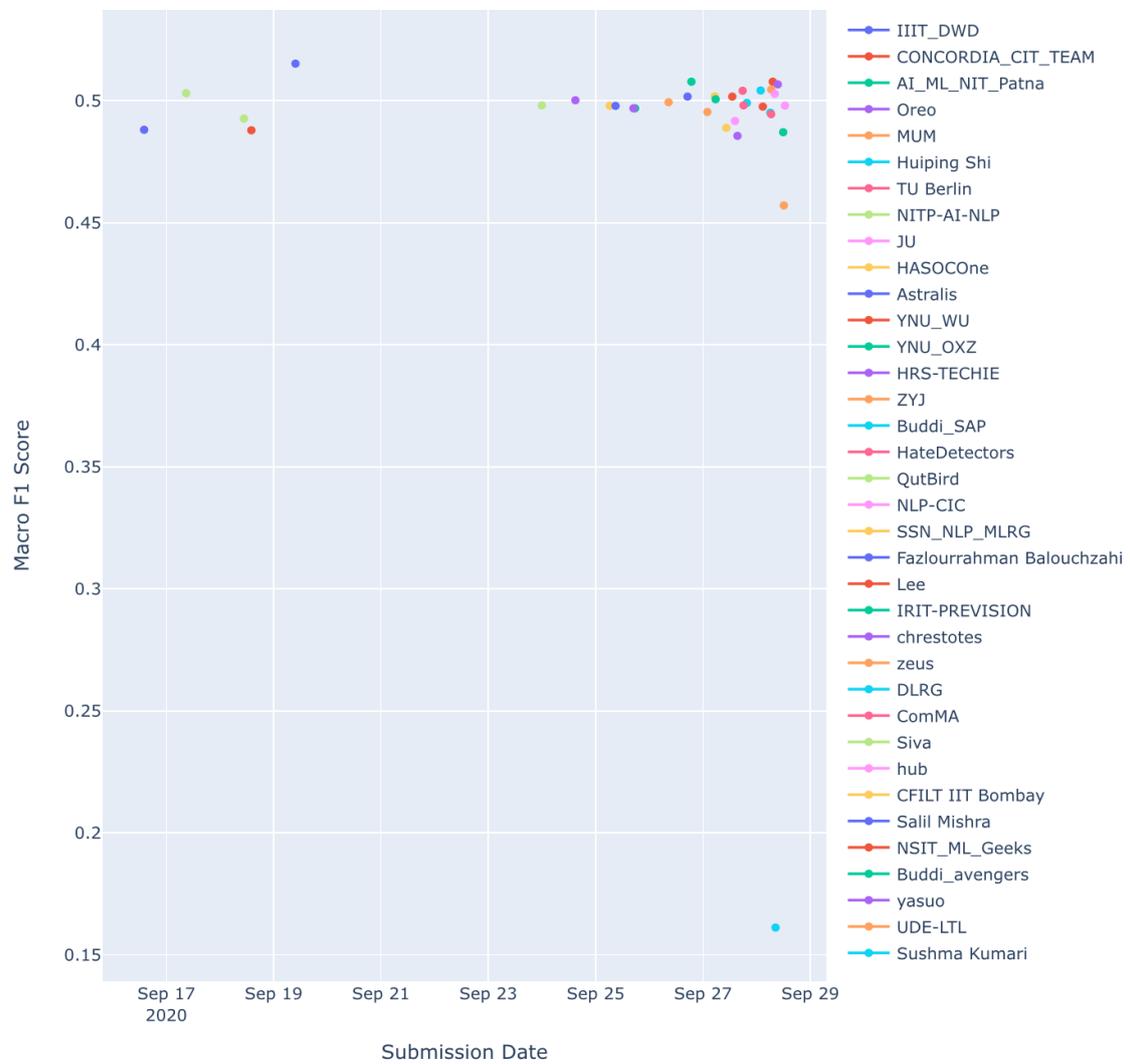

**Figure 9:** English Task A

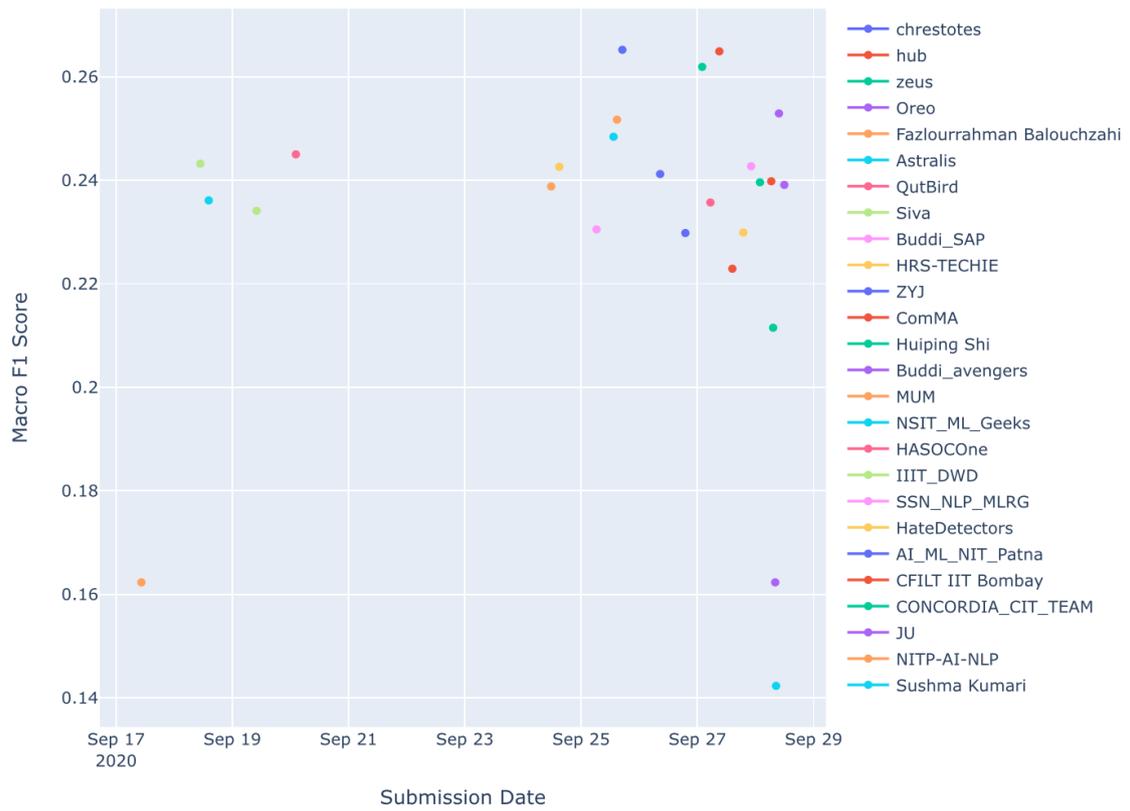

**Figure 10:** English Task B

| Team | English | | German | | Hindi | |
|---|---|---|---|---|---|---|
| Dataset | Number of Submission | | | | | |
| | Task A | Task B | Task A | Task B | Task A | Task B |
| IIIT_DWD | 1 | 1 | 1 | 1 | 1 | 1 |
| CONCORDIA_CIT_TEAM | 1 | 1 | 1 | 1 | 1 | 1 |
| AI_ML_NIT_Patna | 1 | 1 | 1 | 1 | 1 | 1 |
| Oreo | 6 | 2 | 1 | 1 | 2 | 1 |
| MUM | 3 | 2 | 1 | 2 | 2 | 3 |
| Huiping Shi | 6 | 5 | 2 | 1 | - | - |
| TU Berlin | 1 | - | 1 | - | 1 | - |
| NITP-AI-NLP | 1 | 1 | 1 | 1 | 1 | 1 |
| JU | 2 | 2 | 1 | 1 | 1 | 1 |
| HASOCOne | 6 | 7 | 4 | 4 | 1 | 2 |
| Astralis | 2 | 1 | 1 | 1 | 1 | 1 |
| YNU_WU | 3 | - | - | - | - | - |
| YNU_OXZ | 2 | - | 3 | - | 2 | - |
| HRS-TECHIE | 6 | 4 | - | - | - | - |
| ZYJ | 2 | 1 | - | - | - | - |
| Buddi_SAP | 2 | 2 | - | - | - | - |
| HateDetectors | 2 | 2 | - | - | 2 | 1 |
| QutBird | 8 | 1 | - | - | 2 | - |
| NLP-CIC | 2 | - | - | - | - | - |
| SSN_NLP_MLRG | 1 | 1 | 2 | 1 | 2 | 2 |
| Fazlourrahman Balouchzahi | 4 | 4 | 2 | - | 3 | - |
| Lee | 1 | - | - | - | - | - |
| IRIT-PREVISION | 2 | - | - | - | - | - |
| chrestotes | 1 | 2 | - | - | - | - |
| zeus | 1 | 1 | - | - | - | - |
| DLRG | 4 | - | 2 | - | 2 | - |
| ComMA | 4 | 4 | 4 | 4 | 4 | 5 |
| Siva | 1 | 1 | 1 | 1 | 1 | 1 |
| hub | 2 | 1 | 2 | 1 | - | - |
| CFILT IIT Bombay | 1 | 1 | 1 | 1 | 1 | 1 |
| Salil Mishra | 1 | - | - | - | - | - |
| NSIT_ML_Geeks | 1 | 1 | 2 | 2 | 1 | 1 |
| Buddi_avengers | 1 | 1 | 2 | 2 | - | - |
| yasuo | 2 | - | 1 | - | - | - |
| UDE-LTL | 2 | - | - | - | - | - |
| Sushma Kumari | 1 | 1 | 1 | 1 | 1 | 1 |
| simon | - | - | 1 | 1 | - | - |
| IRLab@IITVaranasi | - | - | 1 | - | 2 | - |
| YUN111 | - | - | - | - | 1 | 1 |
| LoneWolf | - | - | - | - | 2 | - |
| # Teams = 40 | 88 | 50 | 40 | 28 | 38 | 25 |

Table 12
Statistics of Team-level Submission. The shaded region in yellow colour represents the team participated in all of the language sub-tasks.

## 6. Discussion and Interpretation

The top teams are are close together. This shows that despite a variety of approaches that was used, no advantage of a particular technology was identified. Most participants used deep learning models and in particular transformer based architectures were popular. Variants of BERT like ALBERT were used much. The best systems for the tasks have applied the following methodology. The best submission for Hindi used a CNN with fastText embeddings as input [20]. The best performance for German was achieved using fine-tuned versions of BERT, DistilBERT and RoBERTa [25]. The best result for English is based on a LSTM which used GloVe embeddings as input [31]. Very heterogeneous approaches led to the best for the respective languages.

For Task B, the best systems reached 0.29 for German, 0.33 [29] for Hindi and 0.26 for English [36]. The fine-grained classification turned out to be a big challenge.

## 7. Conclusion and Outlook

The results of HASOC 2020 have shown that hate speech identification remains a difficult challenge. The performance measures for the test set in HASOC 2020 have been considerably lower than those for HASOC 2019. This is likely an effect of the different data sampling method. Despite the fact that the method is close to realistic proceedings at a platform, it has led to much profane content. The best results are achieved with state of the art transformer models and its variants like ALBERT. The differences between the results for the three languages are small. This seems to indicate that pre-trained deep learning models have the potential to deliver good performance even for languages with little traditional resources. The organizers hope that the data will be used for further research related to hate speech. Apart from classification, topic modelling, analysis of the reliability of the evaluation and failure analysis seem promising areas of research.

For future evaluations, the analysis of language might need to be supplemented by an analysis of visual material posted in social media. Often, offensive intent can only be seen when considering both text and, e.g. image content [38]. Many hateful tweets are also shared as misinformation. We can also have a look at the hateful tweets, which are spread as misinformation[39, 40, 41].

The identification of offensive content still leaves the social questions unanswered: How to react? Different approaches have been proposed; they reach from deletion [42] to labeling [43] and to counter speech by either bots [44] or humans [45]. Societies need to find strategies adequate for their specific demands. We could also use a different kind of algorithm like a spiking neural network to improve the performance of hate speech detection using temporal and non-temporal features. [46, 47].

## 8. Acknowledgement

We thank all participants for their submissions and their valuable work. We thank all the jurors who labelled the tweets in a short period of time. We also thank the FIRE organisers for their support in organising the track.